\documentclass[sn-apa]{sn-jnl}% APA Reference Style 
\usepackage{graphicx}%
\usepackage{multirow}%
\usepackage{amsmath,amssymb,amsfonts}%
\usepackage{amsthm}%
\usepackage{mathrsfs}%
\usepackage[title]{appendix}%
\usepackage{xcolor}%
\usepackage{textcomp}%
\usepackage{manyfoot}%
\usepackage{booktabs}%
\usepackage{algorithm}%
\usepackage{algorithmicx}%
\usepackage{algpseudocode}%
\usepackage{listings}%
\usepackage{appendix}
\usepackage[T1]{fontenc}

\theoremstyle{thmstyleone}%
\theoremstyle{thmstyletwo}%

\theoremstyle{thmstylethree}%

\begin{document}

\title[Article Title]{A Strategic Framework for Optimal Decisions in Football 1-vs-1 Shot-Taking Situations: An Integrated Approach of Machine Learning, Theory-Based Modeling, and Game Theory}


\author[1]{\fnm{Calvin C. K.} \sur{Yeung}}\email{yeung.chikwong@g.sp.m.is.nagoya-u.ac.jp}
% ORCID 0000-0002-1895-1344
% \author[2]{\fnm{Tony} \sur{Sit}}\email{tony.sit@cuhk.edu.hk}
% % ORCID 0000-0001-5006-1838
\author*[1,2,3]{\fnm{Keisuke} \sur{Fujii}}\email{fujii@i.nagoya-u.ac.jp}
% ORCID 0000-0001-5487-4297

\affil[1]{\orgdiv{Graduate School of Informatics}, \orgname{Nagoya University}, \orgaddress{\city{Nagoya}, \country{Japan}}}

% \affil[2]{\orgdiv{Department of Statistics}, \orgname{The Chinese University of Hong Kong}, \orgaddress{\city{Hong Kong SAR}, \country{China}}}

\affil[2]{\orgdiv{Center for Advanced Intelligence Project}, \orgname{RIKEN}, \orgaddress{\city{Osaka}, \country{Japan}}}

\affil[3]{\orgdiv{PRESTO}, \orgname{Japan Science and Technology Agency}, \orgaddress{\city{Saitama}, \country{Japan}}}


\abstract{
Complex interactions between two opposing agents frequently occur in domains of machine learning, game theory, and other application domains. Quantitatively analyzing the strategies involved can provide an objective basis for decision-making. One such critical scenario is shot-taking in football, where decisions, such as whether the attacker should shoot or pass the ball and whether the defender should attempt to block the shot, play a crucial role in the outcome of the game. However, there are currently no effective data-driven and/or theory-based approaches to analyzing such situations. To address this issue, we proposed a novel framework to analyze such scenarios based on game theory, where we estimate the expected payoff with machine learning (ML) models, and additional features for ML models were extracted with a theory-based shot block model. Conventionally, successes or failures (1 or 0) are used as payoffs, while a success shot (goal) is extremely rare in football. Therefore, we proposed the Expected Probability of Shot On Target (xSOT) metric to evaluate players' actions even if the shot results in no goal; this allows for effective differentiation and comparison between different shots and even enables counterfactual shot situation analysis.
In our experiments, we have validated the framework by comparing it with baseline and ablated models. Furthermore, we have observed a high correlation between the xSOT and existing metrics. This alignment of information suggests that xSOT provides valuable insights. Lastly, as an illustration, we studied optimal strategies in the World Cup 2022 and analyzed a shot situation in EURO 2020.
}

% Why did the study need to be done?
% What did you do?
% What did you find?
% How will this study advance the field?
% Paragraph structure
% Brief background (no more than 2 sentence) (why this study need to be done)
% Use signposting (eg, however) to identify research question (why this study need to be done)
% Method/aim (what have you done)
% Result (what you find)
% Conclusion (How it advances the field)




\maketitle

\section{Introduction}
\label{sec:introduction}
    Understanding the interaction between agents, involving the dynamic exchange, communication, and coordination among them, is a fundamental issue in our social activities. It positively affects decision-making and teamwork, providing valuable insights into both the interactions and the agents involved; this holds great significance in various fields, including artificial intelligence \citep{tuyls2021game}, robotics \citep{liu2021motor}, game theory \citep{palacios2003professionals}, and social sciences \citep{axelrod1981evolution}.
    To study such topics, an investigation of the interaction between agents, which occurs between two entities or individuals under a specific context, is required. 

    In team sports, the interaction between two agents refers to the strategies, movements, and decisions made by two opposing players or teams. 
    It includes factors such as positioning, communication, and cooperation between the players to outwit, outmaneuver, or counter each other's actions. 
    The interaction between two agents in sports greatly influences the flow of the game, the outcome of specific plays, and even the final outcome.

    Football has been one of the most influential team sports \citep{li2020analysis,zhang2022sports,yeung2023transformer}, where the outcome of the game is greatly influenced by the critical event of taking a shot. 
    However, despite its significance, the existing literature lacks effective, data-driven, and theory-based methods to comprehensively understand and analyze the interaction strategy between the shooter and the defender. 
    
\begin{figure}[]
    \centering
    \includegraphics[width=0.6\textwidth]{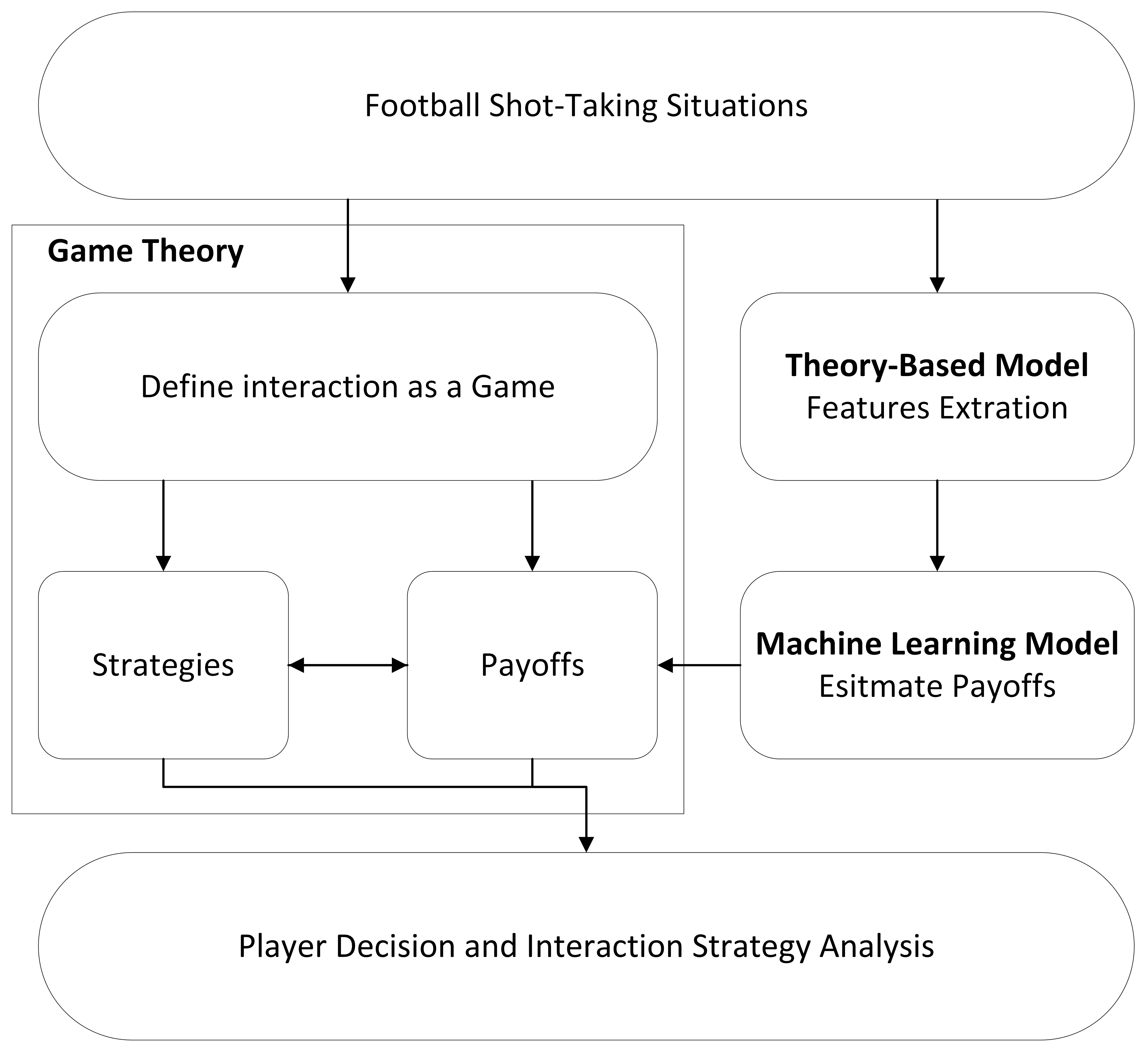}
    \caption{Refined concept of the proposed approach. The assessment of strategies in this study utilizes game theory as its foundation. Machine learning models are employed to estimate the payoffs and value associated with players’ actions. Additionally, a theory-based model is utilized to extract further informative features, enhancing the analysis process.}
    \label{fig:xsot_concept}
    \end{figure}

    To address these issues, we propose a novel approach for gaining deeper insights into the interaction strategy between the shooter and the closest defender, as well as evaluating the shooter's decisions in each specific situation. The method employs game theory, which has been conventionally adopted for interaction strategy analysis, to determine the best interaction strategy for both opposing players. Nevertheless, since a goal is rare in football, it would be hard to determine the values of players' actions (payoff for each strategy under the game theory). Therefore, we employ Machine Learning models to estimate the values of players' actions. In addition, we proposed a novel theory-based model shot block model to extract more informative features for the machine learning model. Finally, Figure \ref{fig:xsot_concept} depicts the concept of the proposed approach.

    For a defender, blocking shots from the shooter might seem to be intuitive.
    However, there might be more effective strategies, for example, not blocking the shot.
    Recently, a professional football coach pointed out that, not attempting to block shots from long-range might be a smart choice\footnote{``Are Liverpool breaking a sacred defensive code?'' (Summersell, 2022), 
    \url{https://medium.com/@chris.summersell/are-liverpool-breaking-a-sacred-defensive-code-8c5f806a4c41}.}\footnote{Example from the above article: \url{https://streamable.com/f88fs8}.}.
    
    There are multiple benefits to not blocking long shots: First, tempting the offensive team to shoot from lower-value locations (i.e., locations that are unlikely to score from) rather than seeking a better opportunity to attack the goal; Second, allowing the goalkeeper a clearer line of sight to control and predict the football; Last, when a block is made, it is likely to end up in a second-ball situation and in areas that cannot be predicted. However, as shown in this study, this might not be the optimal strategy that achieves Nash equilibrium \citep{nash1951non}.
    % With simple statistics and qualitative analysis, it shows evidence that Liverpool doesn't often block long shot\footnotemark[1]. This indicates that there could be alternative strategies for defenders.

    To summarize, this research aims to analyze two-agent interactions in football shot-taking. The contributions of this research are as follows: 
    1) A novel framework and metrics that could analyze attacker and defender strategy, identify the optimal decision and evaluate player action value;
    2) Proposed an effective approach to integrate the machine learning model, theory-based model, and game theory to analyze opposing agents' interaction under complex situations, typically in sports;
    3) With openly accessible data, we verified our proposed framework and metrics by comparing them with baseline models, ablated models, and existing metrics. Moreover, examples of strategy analysis with World Cup 2022 and in-depth shot-taking situation analysis with EURO 2020 were included.

\section{Related work}
\label{sec:related_work}
    In this section, the existing literature on analyzing strategy and decision will be addressed. Subsequently, the evaluation of the actions and decisions of the players will be discussed.
    
    \subsection{Strategy and Decision Analysis in Football}
    In the domain of reinforcement learning, simulated football environments have been extensively utilized for studying football-playing strategies. They can be broadly categorized into two types: humans and robots. Firstly, environments developed based on real-world football. These include the Gameplay Football simulator\footnote{(Schuiling,2017), \url{https://github.com/BazkieBumpercar/GameplayFootball}.}, the older version of DeepMind MuJoCo Multi-Agent Soccer Environment \citep{liu2019emergent}, and Google Research Football \citep{kurach2020google}. Secondly, environments specifically designed for developing football-playing strategies for robots and humanoids. Such as the Robo Cup Soccer Simulator \citep{kitano1997robocup,kitano1998robocup}, and the
    DeepMind MuJoCo Multi-Agent Soccer Environment \citep{haarnoja2023learning}.
    Nonetheless, the strategies developed within these simulated environments have not been verified in real-world football scenarios.
    
    Several efforts have been made to bridge the gap between simulated and real-world football environments. For instance, comparing strategy in simulated and real-world football environments via social network and correlation analysis \citep{scott2021does}, applying the strategy developed in a simulated robot environment to real-world robot zero-shot \citep{haarnoja2023learning}, and utilizing real-world data to develop the strategy in the simulated environment \citep{fujii2023adaptive}. Nevertheless, the interaction strategy of both opposing agents involved has not been the focal point.

    Conversely, with real-world data, reinforcement learning techniques, such as Markov Decision Processes (MDPs), have been utilized to identify actions that maximize rewards during a possession period \citep{rahimian2022beyond,rahimian2023towards}. The rewards are based on expected goals (xG) \citep{macdonald2012expected,eggels2016expected}. Advancements have been made by expanding the action space to include shooting and movement options, as well as considering different pitch locations, enabling a more detailed optimal action analysis \citep{van2021analyzing,van2021learning}. 
    However, reinforcement learning focuses on learning agents maximizing rewards (mainly goals), which makes it usually difficult to analyze the decision-making precisely. On the other hand, game theory emphasizes modeling and considering the strategies of both opposing agents involved. Hence, game theory is applied in this study.

    In the domain of game theory, penalty kicks have been the primary focus of interaction strategies. Both goalkeepers' and shooters' optimized strategies have been analyzed using statistical methods and game theory \citep{palacios2003professionals}. Building upon previous work, the inclusion of a clustering method to differentiate between player roles has allowed for a more in-depth and player role-driven analysis of strategies with game theory \citep{tuyls2021game}.
    
    However, it is imperative to note that penalty kicks are rare events in a match and are independent of other outfield players or previous game states. Meanwhile, shots are significantly more frequent, hold equal importance, and involve more complex decision-making processes. Consequently, analyzing shots provides deeper insights into the game dynamics.

    \subsection{Player Action Evaluation}
    Goals have conventionally been employed, whether as rewards for reinforcement learning or for evaluating player and team performance. However, a notable drawback of using goals is their rarity, resulting in a scenario where the value or reward associated with them is often zero. This scarcity poses challenges for reinforcement learning algorithms because it becomes difficult to learn from and generalize such infrequent events. Similarly, when evaluating players, the limited occurrence of goals may hinder the accurate assessment of players' contributions. To address this limitation, researchers have employed machine learning and theory-based approaches. The approaches aim to model the expected probability of success of specific actions, using them as the value of player action either directly or indirectly.

    In the domain of machine learning, the Expected goal (xG) \citep{macdonald2012expected,eggels2016expected} and Expected Goal Value (EGV) \citep{lucey2015quality} have been proposed to estimate the expected probability of a goal. The metrics Valuing Actions by Estimating Probabilities (VAEP) \citep{decroos2019actions} and Goal Impact Metric \citep{liu2018deep,liu2020deep} have extended the idea, where the player action value is the change of expected probability in scoring and/or conceding between actions, as well as variants of VAEP that focus on defense, VDEP \citep{toda2022evaluation} and GVDEP \citep{umemoto2022location}. 
    
    Beyond evaluating players with goals, some metrics evaluate the expected probability of assist Expected Assist (xA)\footnote{xA, \url{https://www.statsperform.com/opta-analytics/}.}, the expected probability that the possession will lead to an attack, Possession Utilization score (poss-util) \citep{simpson2022seq2event} and the advanced one \citep{yeung2023transformer}.

    In the domain of theory-based models, important elements are often decomposed based on domain knowledge. Each element is then modeled using theories of statistics and physics. One such model is the Expected Threat (xT) \footnote{xT, \url{https://karun.in/blog/expected-threat.html}.}, which quantifies the opportunities created by a player. xT breaks down the threat into probabilities of movement, shot, goal, and transition of zones, estimating these probabilities using historical statistics.
    
    The Dangerousity (DA) metric \citep{link2016real} estimates the probability of a player scoring a goal while in possession of the ball. It considers factors such as zone, control, pressure, and density (chance), and models each of these elements using theories of statistics and physics. Based on the DA, the Off-Ball Scoring Opportunity (OBSO) \citep{spearman2018beyond, spearman2017physics} models the probability of an off-ball player scoring. Furthermore, researchers have integrated both machine learning and theory-based approaches. The C-OBSO \citep{teranishi2023evaluation} proposed a modified score model to consider the defenders' locations.

    Nonetheless, most metrics have focused on the success or failure of actions, such as a shot, pass, or cross. However, the outcome of each action can have multiple outcomes; for example, the outcome of a shot can be categorized as shot on target, shot off target, or shot blocked. 

    %Where the ML model is used and why
    Therefore, in this study, we not only considered various factors that influence the outcome of a shot, but we also decomposed the shot outcome and utilized machine learning models to predict each outcome of a shot. This approach can provide us with a deeper understanding of the game and enable more complex analyses.

    %Where the theory-based model is used and why
    Furthermore, we utilized an improved theory-based shot block model to estimate the probability of a shot being blocked for the shot block outcome, considering both the shooter and defender features. Subsequently, this shot block probability was incorporated as a feature in the machine learning model for the shot block. Our findings indicated that this approach outperformed directly fitting defender features into the machine learning model. Further details are mentioned in Section \ref{sec:rule_based_model}.

\section{Methods}

\begin{figure}[]
    \centering
    \includegraphics[width=\textwidth]{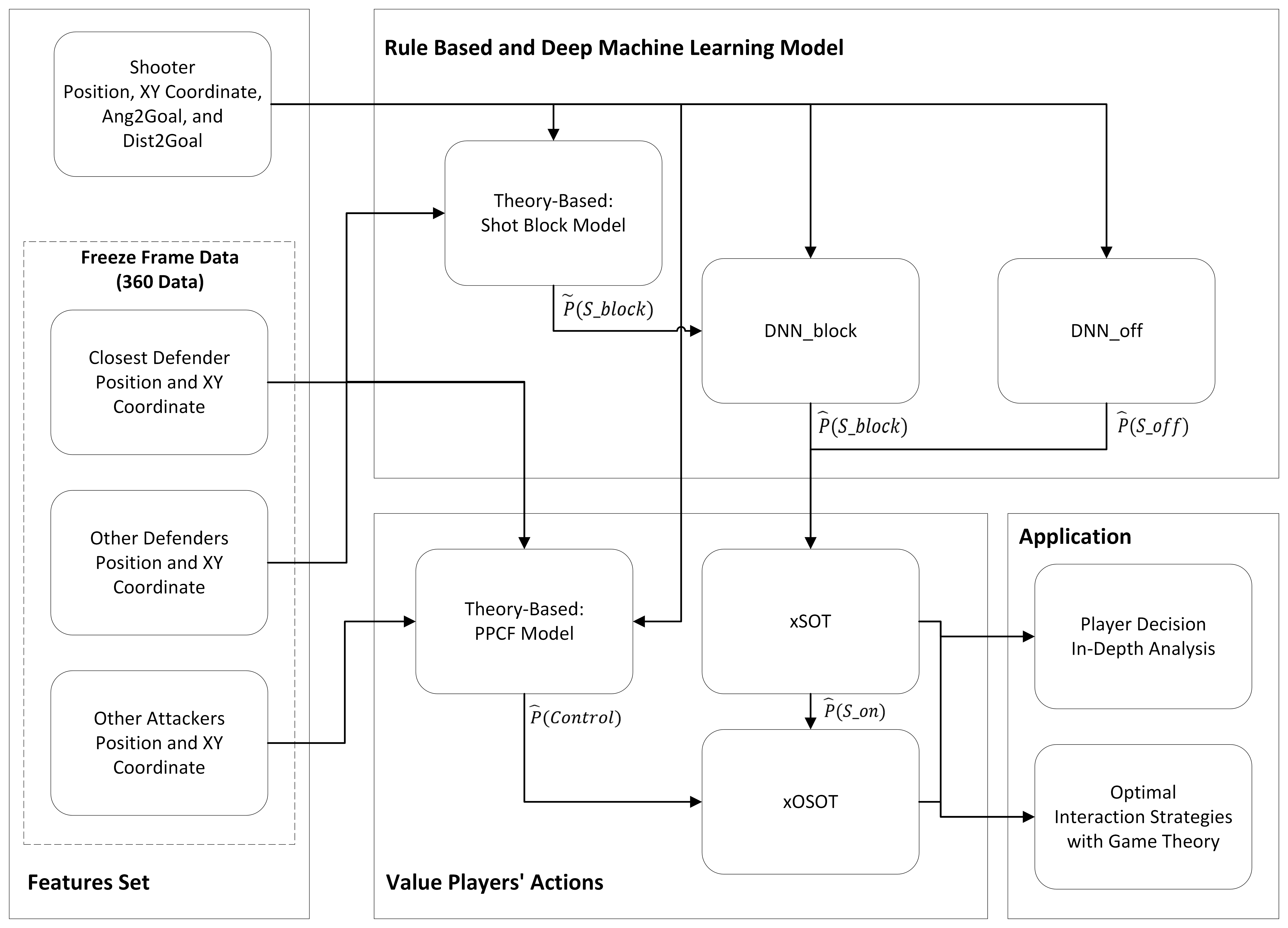}
    \caption{Flow chart of the proposed framework. In terms of game theory payoffs, if the closest defender chooses not to block, the xSOT and xOSOT are calculated without incorporating the closest defender features. For xOSOT, the xSOT$_a$ is calculated using the same method, but instead of the shooter, it was replaced with the other attacker $a$.   
    }
        \label{fig:xsot_framwork}
    \end{figure}

    This section explains how the interaction between the shooter and the closest defender can be formulated as a game, along with the modeling of the relative payoffs. The framework commences with a feature set derived from event and freeze frame data. Subsequently, a combination of shot-blocking theory-based and deep learning (DNN) methods is employed to estimate the value of players' actions, specifically the probability of action outcomes. Finally, the determined value of a player's action is utilized to conduct a comprehensive analysis of their decision-making process and optimize interaction strategies using game theory. Figure \ref{fig:xsot_framwork} depicts the details of the proposed framework.

    \subsection{Define Interaction as a Static Game with Game Theory}
    The initial step in considering the interaction between the shooter and closest defender as a static game is to define the strategy profile $S_i$ for agent $i$ and the corresponding payoff. The strategy profiles for the shooter and closest defender are defined as follows:
    \begin{equation}
    \label{eq1}
    S_{shooter}\in\{\text{Shoot, Pass}\}, \ \ \ \ 
    S_{defender}\in\{\text{Blocking, Not Blocking}\}.
    \end{equation}
    The shooter of the attacking team has two options: either to shoot at their current location or to pass the ball to other players in the attacking team, allowing them to shoot at their respective locations. On the other hand, the closest defender also has two choices: attempting to block the shooter's shot or applying Liverpool's strategy, which involves not blocking the shot and potentially gaining certain benefits (as mentioned in Section \ref{sec:introduction}).
    
    Furthermore, the payoffs for each combination of strategies depend on the current state of the football match game. The ultimate goal of every player is to win the match. Traditionally, the probability of scoring goals has been used as the payoff or reward. However, scoring goals is a rare event that involves randomness, and expecting players to score on every shot they take is unrealistic. Therefore, we focus on the minimum requirement of taking a shot, which is shot on target. We summarize the outcome event space of taking a shot as follows: 
    \begin{equation}
    \label{eq2}
    \text{Shot Outcome} \in \{\text{Shot On Target, Shot Off Target, Shot Block}\}.
    \end{equation}
    The detailed explanation can be found in Section \ref{sec:shot_outcome_appendix}.

    For the shooter, we define the payoff for shooting as the Expected Probability of Shot On Target (xSOT), representing the likelihood of the shot being on target. Conversely, the payoff for passing is defined as the Expected Probability of Off-Ball Player Shot On Target (xOSOT), indicating the probability of a successful shot from another player on the attacking team. As for the closest defender, their payoff is the negative of the shooter's payoff. When the closest defender chooses not to block, the xSOT and xOSOT are calculated without considering the closest defender. The payoffs for the shooter and defender are summarized in Table \ref{tab:game_theory}.
    \begin{table}
    \caption{Game theory payoff table.}\label{tab:game_theory}
    \begin{tabular}{cccc}
    \hline
                              &                            & \multicolumn{2}{c}{Defender}                                              \\
                              &                            & Blocking                            & Not Blocking                         \\ 
                              \hline

    \multirow{2}{*}{Shooter} & \multicolumn{1}{c}{Shoot} & \multicolumn{1}{c}{xSOT,-xSOT} & \multicolumn{1}{c}{xSOT,-xSOT} \\  
                              & \multicolumn{1}{c}{Pass}  & \multicolumn{1}{c}{xOSOT,-xOSOT} & \multicolumn{1}{c}{xOSOT,-xOSOT} \\  
                              \hline
    \end{tabular}

    \footnotetext{The left value indicates the payoffs for the shooter and the right value indicates the payoffs for the closest defender.}
    \end{table}

    %Nash equilibrium
    Moreover, finding the optimal interaction strategy for both the shooter and closest defender is equivalent to identifying the Nash equilibrium. The Nash equilibrium is defined as follows \citep{nash1951non, tuyls2021game, palacios2003professionals, tadelis2013game}:
    
    Let $s^*=(s_{i}^*,s_{-i}^*),\ s_i \in S_i$ be a strategy profile with a strategy for each agent, where $s_{-i}$ denote the strategy for agents other than agent $i$ and $i \in \{\text{attacker, defender}\}$. 
    Let $u_i(s_{i},s_{-i}^*)$ be the payoff for agent $i$.
    The strategy profile $s^*$ is a Nash equilibrium if and only if,
    \begin{equation}
    \label{eq3}
    \mathbb{E}[u_i(s_{i}^*,s_{-i}^*)]\geq \mathbb{E}[u_i(s_{i}',s_{-i}^*)]\  \ \ \ \forall\ s_i'\in S_i\ , i \in \{\text{attacker, defender}\}
    \end{equation}
    %Assumption
    Lastly, the following assumptions are made for the game:
    
    \begin{itemize}
    \item Relational decision maker: Each agent will make rational decisions by choosing the best strategy available to them \citep{tadelis2013game}.
    \item Complete information: All agents possess complete knowledge of the game, and this knowledge is common among all participants \citep{tadelis2013game}.
    \item Static one-stage game: The nature of the game, whether static or dynamic, is discussed in Section \ref{sec:data_independence_test}. However, for the current analysis, we assume a static one-stage game due to the unavailability of players' velocity and other detailed data required to model and analyze their future movements.
    \end{itemize}
    
    \subsection{Calulate xSOT with Machine Learning Models}
    \label{sec:xsot}
    When modeling the xSOT, we consider all possible outcomes of a shot, including shot on target ($S_{\text{on}}$), shot off target ($S_{\text{off}}$), and shot block ($S_{\text{block}}$) (details explanation in Section \ref{sec:shot_outcome_appendix}). Since the set $\{ S_{\text{on}},S_{\text{off}},S_{\text{block}}\}$ is taken as the sample space of shot outcomes, we can model the xSOT using the law of total probability. Consequently, the xSOT can be represented by the following equations:
    \begin{align}
    \label{eq4}
    xSOT=\mathbb{E}[P(S_{\text{on}})]&=\mathbb{E}[1-min(P(S_{\text{off}})+P(S_{\text{block}}),1)],\nonumber\\
    \hat{P}(S_{\text{off}})&=DNN_{\text{off}}(\vec{x}_{\text{off}},y_{\text{off}}),\\
    \hat{P}(S_{\text{block}})&=DNN_{\text{block}}(\vec{x}_{\text{block}},y_{\text{block}}),\nonumber
    \end{align}
    
    \noindent where the $P(S_{\text{off}})$ and $P(S_{\text{block}})$ are estimated with a Deep Neural Network (DNN) (also known as MLP: multilayer perceptron) for classification respectively, and trained with cross-entropy loss (CEL). The hyperparameters for the DNN  and the optimized values are listed in Section \ref{sec:hyperparameters}.
    
    Moreover, $\vec{x},\ y$ are the input features vector and target features for the DNN model, respectively. Both $\vec{x}_{\text{off}}$ and $\vec{x}_{\text{block}}$ consist of the following basic shooter features, where the first three features are the event data and adhere to the definition from StatsBomb
    \footnote{\label{label:statsbomb1}More Details can be found at, \url{https://github.com/statsbomb/open-data/blob/master/doc/Open\%20Data\%20Events\%20v4.0.0.pdf}.}:
    \begin{itemize}
        \item player role: The role of the player, for instance, center forward, center back, goalkeeper, etc. StatBomb has named this feature as position.
        \item location x: Football pitch coordinate x of the shooter. Represent the length dimension of the football pitch ranging from 0 to 120. 
        \item location y: Football pitch coordinate y of the shooter. Represent the width dimension of the football pitch ranging from 0 to 80. 
        \item Dist2Goal: Distance from the shooter to the middle of the goal line. Calculated with Equation \ref{eq11}.
        \item Ang2Goal: Absolute angle from the shooter to the middle of the goal line. Calculated with Equation \ref{eq11}.
    \end{itemize}
    For $\vec{x}_{\text{block}}$, in addition to the previously mentioned features, we incorporate the location and position data of the off-ball players using StatsBomb freeze frame 360 data\textsuperscript{\ref{label:statsbomb2}}; this allows us to create the following additional features:
    \begin{itemize}
        \item Theory-based shot block feature: Shot block probability estimation from a theory-based shot block model (Explained in Section \ref{sec:rule_based_model}) that utilizes the StatsBomb freeze frame 360 data\footnote{\label{label:statsbomb2}More Details can be found at, \url{https://statsbomb.com/what-we-do/soccer-data/360-2/}.}. The freeze frame 360 data includes the position, location x,y of other players on the pitch. However, since the data was collected from a video frame, data for any players that were not in the frame were not included.
    \end{itemize}
    The target variables $y_{\text{off}}$ and $y_{\text{block}}$ will take the value 1 when the outcome is shot off target and shot block, respectively. For all other outcomes, the target variables will have a value of 0. 

    We assess the performance of the $DNN_{\text{block}}$ and $DNN_{\text{off}}$ models by comparing them with baseline models that utilize the same feature set. These baselines include common statistical models, historical percentages derived from the dataset, and ElasticNet \citep{zou2005regularization}. Additionally, we consider tree-boosting models, namely XGBoost \citep{chen2016xgboost} and CatBoost \citep{prokhorenkova2018catboost}, which have been commonly employed in previous studies to model the expected probability of a goal \citep{macdonald2012expected, eggels2016expected} as well as scoring and conceding patterns \citep{decroos2019actions, toda2022evaluation, umemoto2022location}, among others.

    \subsection{Create Additional Feature with Theory-Based Shot Block Model}
    \label{sec:rule_based_model}
    % $$\tilde{P}(S_{\text{block}})=0.4252$$

    \begin{figure}[]
    \centering
    \includegraphics[width=0.9\textwidth]{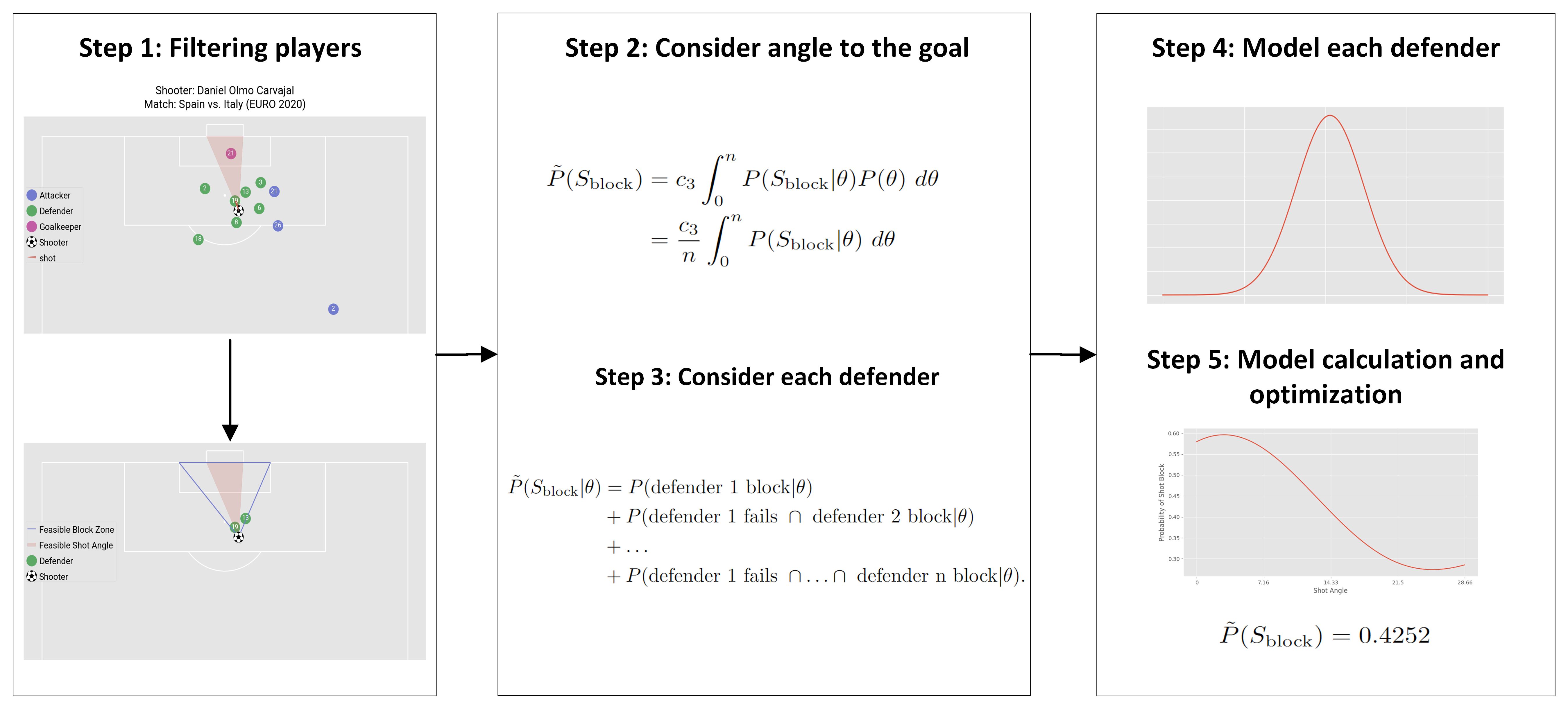}
    \caption{Flow chart of the theory-based shot block model. Each step is explained in detail in this section, Section \ref{sec:rule_based_model}.}
    \label{fig:tb_concept}
    \end{figure}
    
    To better utilize the location and position data of off-ball players (StatsBomb freeze frame 360 data\textsuperscript{\ref{label:statsbomb2}}), we proposed the theory-based shot block model that captures the information in the 360 data. The theory-based model estimates the probability $P(S_{\text{block}})$. The estimated probability was later used as a feature for the deep learning model $DNN_{\text{block}}$ and named the theory-based shot block feature. Figure \ref{fig:tb_concept} depicts the detailed steps of the theory-based shot block model. This theory-based model draws inspiration from the scoring probability model and shot block value in C-OBSO \citep{teranishi2023evaluation}. The main idea of this method is that the farther the ball is from the defender, and the larger the difference in angle, the more difficult it becomes for the defender to block the shot.

    More specifically, the probability of a single defender blocking a shot is modeled using a normal distribution probability density function (PDF).
     Additionally, the shot block probability is calculated by summing a discrete set of angles from the shooter to the goal line, bounded by the goal posts.

    We have made several improvements compared with the C-OBSO approach. Primarily, we excluded the goalkeeper from our considerations, as a saved shot is still counted as being on target. Moreover, we consider the angle to the goal as continuous rather than discrete. This change allows us to achieve a more precise value of the PDF function.

    Moreover, we introduced a more realistic event space, in addition to assuming the probability of each defender is independent, to better reflect the realistic scenario. If one defender has already blocked the shot, other defenders won't be able to block it subsequently. Furthermore, we substituted the normal PDF with a truncated normal PDF. The truncated version restricts the reachable location of the defender rather than extending it infinitely.

    \begin{figure}[]
    \centering
    \includegraphics[width=0.7\textwidth]{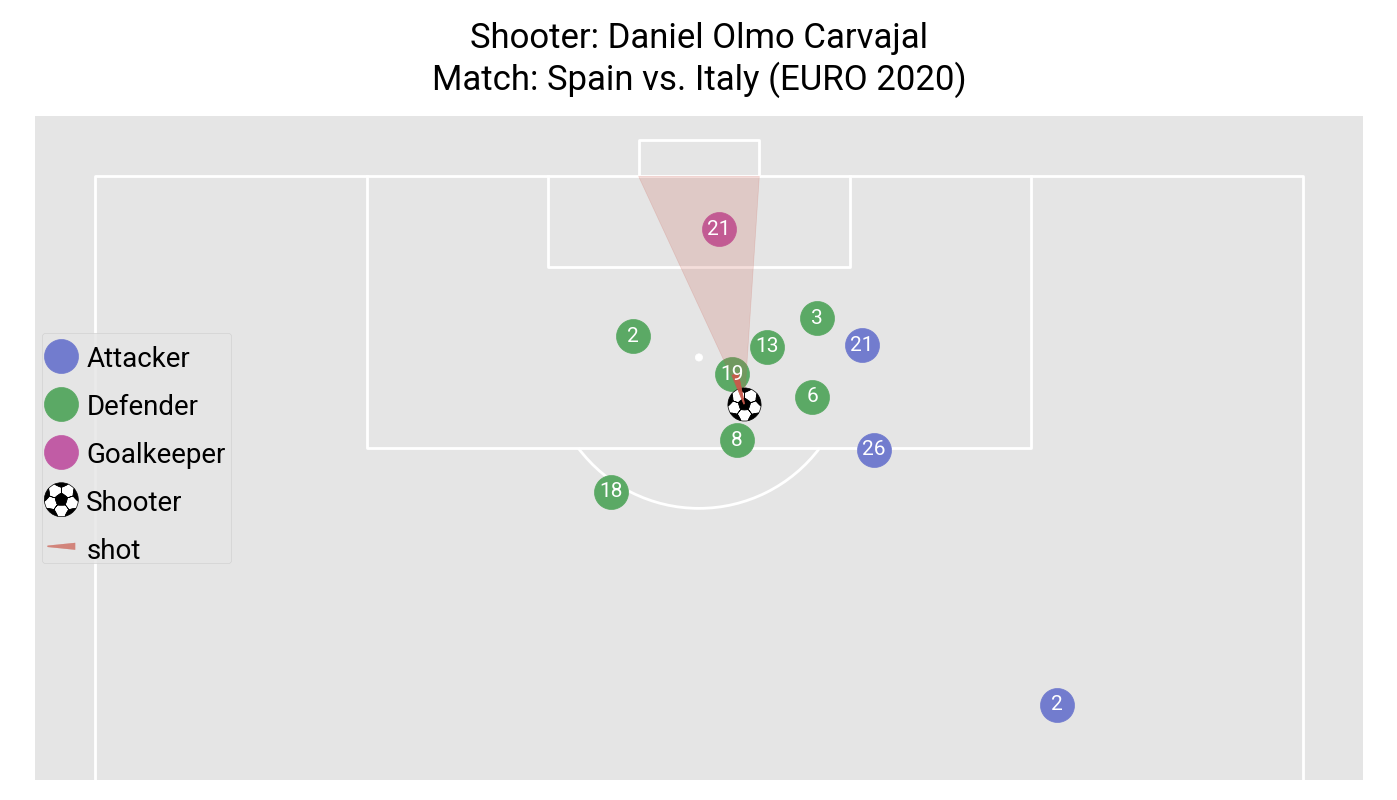}
    \caption{Shot-taking situation example image. The image included all players who appeared in the
    freeze frame from the match Spain vs. Italy, EURO 2020.}
    \label{fig:tb_example}
    \end{figure}
    
    Finally, to ensure the robust and rigorous foundation for our methodology. We explain a specific shot-taking situation as in Figure \ref{fig:tb_example} and provide a statistical theory-based and detailed derivation of the theory-based shot block model as follows:
    %Could summarize the fig in each step (fig3,4,...) into one fig but could be too small 

    \begin{figure}[]
    \centering
    \includegraphics[width=0.7\textwidth]{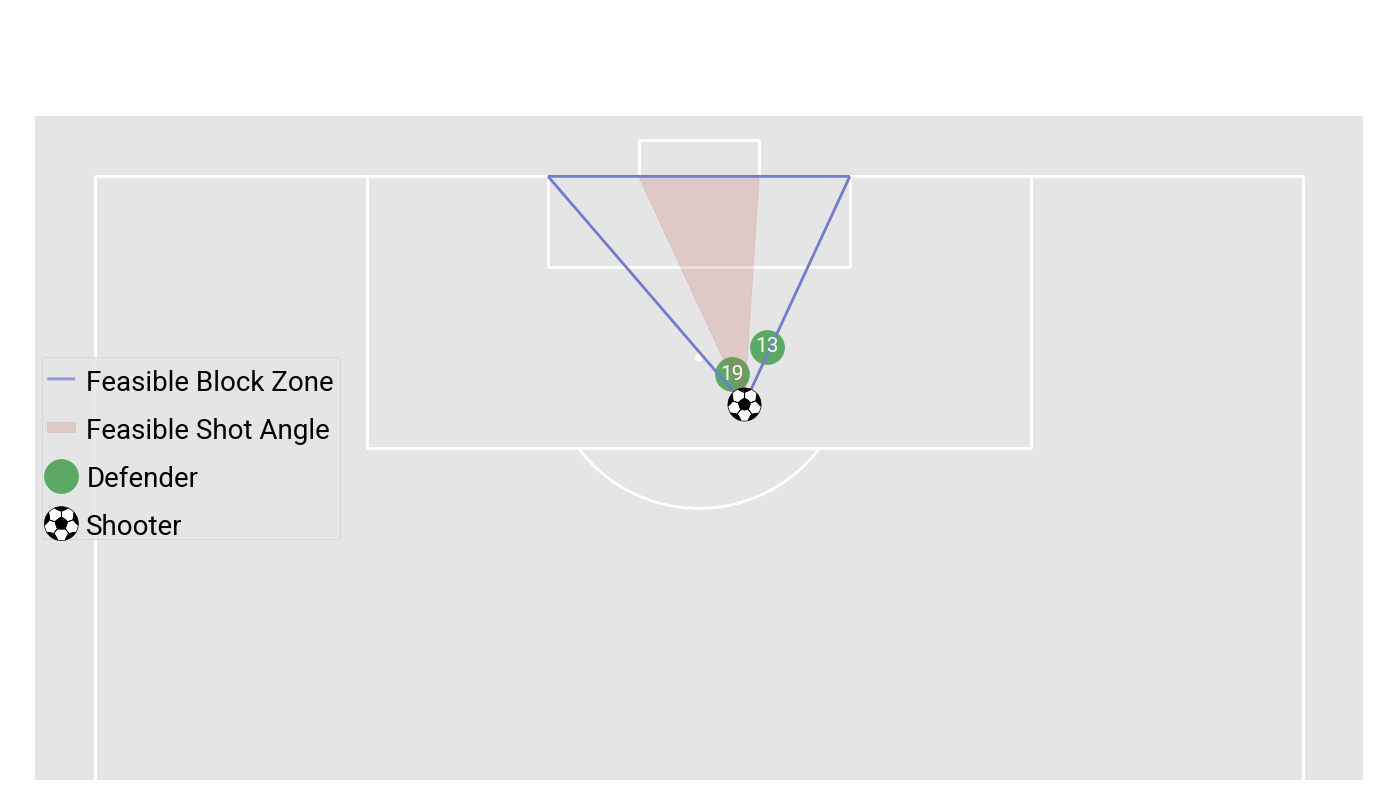}
    \caption{Theory-based shot block model feasible block zone and feasible angle. The shot-taking situation is based on the one in Figure \ref{fig:tb_example}. Defenders inside the feasible block zone bounded by blue lines are considered. The probability of block is calculated based on  shot angles inside red regions. The line from the left goalpost to the shooter is considered as shot angle 0, and from the right goalpost to the shooter is considered as shot angle $n$.}
    \label{fig:tb_filter}
    \end{figure}
    
    \textbf{Step 1: Filtering players.} The filtering process begins by excluding the goalkeeper, players on the same team as the shooter, and defenders located outside the feasible block zone bounded by the coordinates of the shooter and the two intersection points between the penalty area line and the goal line (as indicated by the blue lines in Fig \ref{fig:tb_filter}). The defenders that remain after this filtering process are labeled as defender ${1,2,...,n} = D$ and are sorted in ascending order based on their distance from the shooter.

    \textbf{Step 2: Consider angle to the goal.} By applying the law of total probability, the shot block probability can be conditioned on the shot angle $\theta$ that the shooter takes. We assume that the shots are taken in straight lines, and each degree within the feasible angle corresponds to a specific shot angle. The feasible shot angle is defined as the angle formed with the straight line from the shooter to the left goal post (as indicated by the left boundary or the red area in Figure \ref{fig:tb_filter}) and the straight line from the shooter to the right post (as indicated by the right boundary or the red area in Figure \ref{fig:tb_filter}). The total degree, equivalent to shot angle $n$, can be calculated using the law of cosines.
    
    The shot block probability can be represented by the following equation for a continuous shot angle $\theta \in [0,n]$:
    \begin{equation} 
    \label{eq5}
    \begin{split}
    \tilde{P}(S_{\text{block}})&=c_3\int_{0}^n P(S_{\text{block}}|\theta)P(\theta)\ d\theta\\
    &=\frac{c_3}{n}\int_{0}^n P(S_{\text{block}}|\theta)\ d\theta
    \end{split}
    \end{equation}
    where $P(\theta)$ represents the probability of the shooter selecting shot angle $\theta$ to shoot. $\tilde{P}$ were used to differentiate the estimation from the theory-based shot block model and estimation $\hat{P}$ from the DNN model, but they are both the estimated probability of shot block. 
    
    To simplify the analysis, we assume that $P(\theta)$ follows a continuous uniform distribution within the range of $[0, n]$, resulting in the second equation. $P(S_{\text{block}}|\theta)$ denotes the probability that the shot will be blocked by defenders in set $D$, given that shot angle $\theta$ is selected. The term $c_3$ represents a constant.
    
    % \begin{figure}[]
    % \centering
    % \includegraphics[width=0.7\textwidth]{defender_eg.png}
    % \caption{Defender blocking illustration. The red area shows the considered shot angles.}
    % \label{fig:defender_eg}
    % \end{figure}
    \textbf{Step 3: Consider each defender.} After considering each shot angle, we can incorporate each defender in set $D$. It is important to note that only one defender can block the shot. For instance, defender $d$ (e.g., defender 13 in Fig \ref{fig:tb_filter}) will have the opportunity to block the shot only if defender $d-1$ (e.g., defender 19 in Fig \ref{fig:tb_filter}) fails to block it, this allows us to partition the event space and utilize the law of total probability to dissect $P(S_{\text{block}}|\theta)$ as follows: 
    %\begin{equation}
    
    \begin{align}
    \label{eq6}
     \tilde{P}(S_{\text{block}}|\theta) &= P(\text{defender 1 block}|\theta) \nonumber\\
    &+ P(\text{defender 1 fails } \cap \text{ defender 2 block}|\theta) \\
    &+ \ldots \nonumber\\
    &+ P(\text{defender 1 fails } \cap \ldots \cap \text{ defender n block}|\theta). \nonumber
    \end{align}
    %\end{equation}
    If $|D|=0$, indicating that there are no defenders in set $D$, the probability $P(S_{\text{block}}|\theta)$ becomes 0. Consequently, the overall shot block probability $P(S_{\text{block}})$ is also 0.

    Furthermore, we assume that the defenders' probabilities to block the shot are independent. With this assumption, the components of the above equation can be further dissected using the following equation:
    %\begin{equation}
    
    \begin{align}
    \label{eq7}
    & \tilde{P}(\text{defender 1 fails } \cap \ldots \cap \text{ defender $d$ block}|\theta)\nonumber\\
    &=(1-P(\text{defender 1 block}|\theta))(1-P(\text{defender 2 block}|\theta))\\
    &\ \ \ \ \ \raisebox{0.7ex}{...}P(\text{defender $d$ block}|\theta). \nonumber
    \end{align}
    %\end{equation}
    \textbf{Step 4: Model each defender.} We model each defender's expected probability of blocking shots using a truncated normal distribution probability density function (PDF). In this case, we treat the PDF as a simple function without statistical meanings. The use of a truncated normal PDF is preferred because it does not have a tail that extends to infinity, unlike the normal PDF; this ensures that the range of a defender's reach is bounded and helps avoid unrealistic assumptions. The function is as follows:
    \begin{equation}
    \label{eq8}
    \begin{split}
    \begin{gathered}
     \tilde{P} (\text{defender $d$ block}|\theta)=f(x;\mu,\sigma,a,b)=\frac{1}{\sigma}\frac{\varphi(\frac{x-\mu}{\sigma})}{\Phi(\frac{b-\mu}{\sigma})-\Phi(\frac{a-\mu}{\sigma})},\\
    x=\frac{(\theta-\theta_d)}{c_1},\ \mu=0,\ \sigma=c_4+\textit{l}_d*c_2,
    \end{gathered}
    \end{split}
    \end{equation}
    \noindent where $(a, b = -a)$ defines the interval that bounds the function. In the equation, $\theta_d$ represents the shot angle at which defender $d$ is positioned, $\textit{l}_d$ represents the distance between defender $d$ and the shooter, and $c_1$, $c_2$, and $c_4$ are constant terms. Furthermore, $\varphi(x)$ represents the probability density function (PDF) and $\Phi(x)$ represents the cumulative distribution function (CDF) for the standard normal distribution. The equations are given as follows:
    \begin{equation}
    \label{eq9}
    \varphi(x) = \frac{1}{\sqrt{2\pi}} e^{-\frac{1}{2}x^2},\ \ \ \
    \Phi(x) = \frac{1}{2} \left(1 + \text{erf}\left(\frac{x}{\sqrt{2}}\right)\right),
    \end{equation}
    \noindent where, $\text{erf}(x)$ is the error function and is approximated numerically.
    % scaler_1=params[0], scaler_2=params[1], scaler_3=params[2], mean=0, sigma=params[3], a=params[4], b=-params[4])
    %[36.94630071, 12.3578812 ,  0.4997678,   0.15766148, -2.30980703]
    %std=sigma+v['defender_distance']/scaler_2
    %x=(angle_i - angle_p) / scaler1

    \begin{figure}[]
    \centering
    \includegraphics[width=0.8\textwidth]{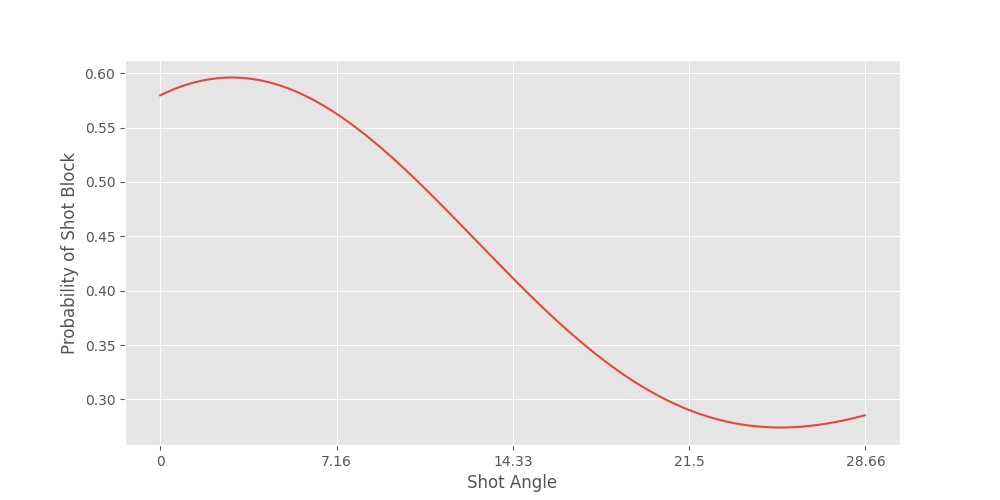}
    \caption{Probability of Shot Block for each feasible shot angle. The shot-taking situation is based on the one in Figure \ref{fig:tb_example}.}
    \label{fig:tb_value}
    \end{figure}

    \textbf{Step 5: Model calculation and optimization}. To ensure computational efficiency, the trapezoidal rule is employed to approximate $P(S_{\text{block}})$ when $|D|>0$ and set it equal to 0 otherwise. For the shot-taking situation in Figure \ref{fig:tb_example}, the probability of shot block at each shot angle is shown in Figure \ref{fig:tb_value}. Various common optimization methods were compared to optimize the parameters and constant terms $c_1, c_2, c_3, c_4, a$. These included iterative-based methods: Powell \citep{powell1964efficient} and Nelder-Mead \citep{nelder1965simplex}, as well as gradient-based methods: CG \citep{hestenes1952methods}, L-BFGS-B \citep{liu1989limited}, and SLSQP \citep{kraft1988software}.

    The results of the optimization process, including the comparison between different optimization methods, can be found in Section \ref{sec:s_block_valid}. Powell was selected as the optimal choice after evaluating the performance of each method due to its superior performance, and the value for the optimized parameters are listed in Supplementary Table \ref{tab:rule_based_opt_param}.
    
    % powell \citep{powell1964efficient}
    % CG  \citep{hestenes1952methods}
    % L-BFGS-B \citep{liu1989limited}
    % SLSQP \citep{kraft1988software}
    % Nelder-Mead \citep{nelder1965simplex}

    \subsection{Calculate xOSOT}
    The xOSOT is calculated by determining the off-ball attacker who has the highest expected probability to shoot on target. The probability of the off-ball attacker being able to control the ball will also be considered since it first requires the shooter to pass the football to the off-ball attacker. This approach modifies the concept of OBSO introduced in \cite{spearman2018beyond}, and further explained in Section \ref{sec:ppcf}. However, in this case, we consider only the off-ball attacker $a \in A$ and the corresponding location that has the highest expected probability to shoot on target, rather than considering all locations on the pitch. The equation for xOSOT is as follows:
    \begin{equation}
    \label{eq10}
    \begin{split}
    \begin{gathered}
    xOSOT=\underset{a\in A}{\mathrm{max}}\ \mathbb{E}[P(S_{\text{on}}\mid \text{Control}_a)*P(\text{Control}_a)], \\
    \hat{P}(S_{\text{on}}\mid \text{Control}_a)=xSOT_a,\\
    \hat{P}(\text{Control}_a)=PPCF_a,
    \end{gathered}
    \end{split}
    \end{equation}
    
    \noindent where $P(S_{\text{on}}\mid \text{Control}_a)$ denotes the probability of a shot on target at the location of the off-ball attacker $a$, given that attacker $a$ has controlled the football. Additionally, $P(\text{Control}_a)$ represents the probability that the ball will be controlled by the off-ball player $a$.

    Furthermore, $xSOT_a$ represents the xSOT calculated with the off-ball attacker $a$. Additionally, $PPCF_a$ denotes the theory-based PPCF model (Potential Pitch Control Field) \citep{spearman2018beyond}, which is calculated from time 0 to $T$, where $T$ is the travel time of the football from the shooter to the off-ball attacker $a$. This is in contrast to the approach in  \citep{spearman2018beyond} where $T\rightarrow \infty$. Considering the finite travel time $T$ is more suitable, as it accounts for the fact that even if the off-ball attacker $a$ gains control of the ball after time $T$, it is unlikely that they can shoot from their current location. 
    
    The PPCF is further explained in Section \ref{sec:ppcf}, providing more details on how it is computed.

\section{Experiments and Results}
    This section aims to verify the xSOT and xOSOT metrics, determine the optimized strategy for the interaction between the shooter and closest defender, and showcase the analysis of each shot-taking situation using xSOT and xOSOT. The code for this study is accessible on GitHub through the following link: \url{https://github.com/calvinyeungck/Football-1-vs-1-Shot-Taking-Situations-Analysis}.
    
    \subsection{Dataset and Preprocessing}
    \textbf{Dataset:} The dataset used for this study was based on the events and freeze frame data from the World Cup 2022 and EURO 2020 tournaments. The football events and freeze frame data were obtained from StatsBomb's free data\textsuperscript{\ref{label:statsbomb2}}, available at \url{https://statsbomb.com/what-we-do/hub/free-data/}.

    The Euro 2020 dataset comprised 51 matches, while the World Cup 2022 dataset included 64 matches. In total, there were 2,575 shot-taking events recorded, with 1,043 shots off target, 850 shots on target, and 682 shots blocked. Additionally, the xG (expected goals) data was sourced from \url{https://footystats.org/international/world-cup/xg}, while the number of goals data was obtained from \url{https://www.mykhel.com/football/fifa-world-cup-2022-team-stats-l4/}.

    \textbf{Preprocessing:} In order to address the limited amount of data, we performed data preprocessing by splitting the dataset into a train set and a test set using the $train\_test\_split()$ function from the Python package sklearn. The ratio was set to 80/20, and the splitting was stratified based on the grouped shot outcome (for more details, refer to Section \ref{sec:shot_outcome_appendix}). For training the DNN and baseline models, we utilized the train set with 5-fold cross-validation, implemented using the $StratifiedKFold()$ function from the sklearn package.
    
    Furthermore, it is important to note that StatsBomb employs a football pitch coordinate system with x ranging from 0 to 120 and y ranging from 0 to 80. However, a professional football pitch typically has a size of 105 meters in length and 68 meters in width. Therefore, we appropriately scaled the xy coordinates. Additionally, we calculated the distance to the goal (Dist2Goal) and angle to the goal (Ang2Goal) features when computing xSOT. The equations for Dist2Goal and Ang2Goal are as follows:
    
    %\begin{equation}
    \begin{align}
    \label{eq11}
    Dist2Goal&=\sqrt{((x-120)*105/120)^2+((y-40)*68/80)^2} \\
    Ang2Goal&=|(arctan(\frac{(40-y)*68/80}{(120-x)*105/120)})| \nonumber
    \end{align}
    %\end{equation}
    Where (x, y) represents the player coordinates, and (120, 40) corresponds to the midpoint of the defending team's goal line in the StatsBomb coordinate system.
    
    \subsection{Models and Framework Validation}
    \label{sec:models_and_framework_validation}

    Here, we validate the effectiveness of using the DNN for modeling the probability of shot off in Section \ref{sec:s_off_valid} and shot block in Section \ref{sec:s_block_valid}. Additionally, we identify the optimal optimization methods for the theory-based shot block model in Section \ref{sec:s_block_valid} and highlight the necessity of the theory-based shot block model in the framework in Section \ref{sec:necessity_tb}.
    
    % For the DNN models, we evaluate the performance of the DNN model by comparing it against baseline models: historical percentage, Elas-
    % ticNet, xGBoost, and CatBoost. These baseline models had been commonly applied to model football events data in previous studies.(see details in Section \ref{sec:xsot})
    
    % Utilizing various optimization methods (see details in Section \ref{sec:rule_based_model}), and considering different feature sets. The evaluation was based on the binary CEL, where a lower CEL indicates better performance. The CEL is a commonly used scoring rule for probability estimation in a 2-class event. 

    \subsubsection{Shot Off Probability Model Validation}
    \label{sec:s_off_valid}

    Beginning with the $DNN_{\text{off}}$ models, we assess their performance by comparing them against baseline models: historical percentage, ElasticNet, xGBoost, and CatBoost, using the same features set as the proposed model. These baseline models have been commonly used to model football event data in previous studies (see details in Section \ref{sec:xsot}). The evaluation will be based on the binary Cross-Entropy Loss (CEL), where a lower CEL indicates better performance. The CEL is a commonly used scoring rule for probability estimation in a 2-class event.

    \begin{table}[]
    \caption{The performance of shot off probability prediction models with machine learning.}
    \label{tab:shotoff}
    \begin{tabular}{lll}
    \toprule
    Model                 & \multicolumn{1}{l}{Avg CEL} & \multicolumn{1}{l}{Std}\\
    \midrule

    \textbf{Ours} ($DNN_{\text{off}}$)& \textbf{0.6696} & 0.0065             \\
    Historical percentage& 0.6749 & 0.0004                     \\
    ElasticNet& 0.6759 & 0.0007                     \\
    XGBoost& 0.9108 & 0.0832                     \\
    CatBoost& 0.9416 & 0.0422   \\
    \botrule
    \end{tabular}
    \footnotetext{The table shows the average (Avg) CEL and standard deviation (Std) on the 5 cross-validations split validation set CEL. The table is ordered with the Avg CEL (the lower the better) in ascending order.}
    \end{table}

    In Table \ref{tab:shotoff}, the performance of the DNN model was compared with other models in estimating the probability of a shot off target. Our model, $DNN_{\text{off}}$, had outperformed all baseline models, and achieved the lowest average CEL of 0.6696. However, it is important to note that $DNN_{\text{off}}$ did not possess an overwhelming advantage compared to other baseline models. More informative features could be engineered in future works.
    % This was primarily due to the fact that, in order to analyze and account for the value created by off-ball attackers, advance shooter features were excluded. 
    % If more features were included, the CEL of $DNN_{\text{off}}$ would be 0.6443. %sigma 0.0042
    % Nonetheless, the performance of $DNN_{\text{off}}$ with a CEL of 0.6696 had demonstrated its effectiveness in providing inference for shot off probability, surpassed the performance of the baseline models.

    % "shot_body_part", "shot_first_time", "shot_technique", "under_pressure
    % Introduce the experiment/figure
    % Discuss the obtained data (Discuss trends, patterns,
    % and/or relationships)
    % Summarize the key founding/ relevance
    % Importance of the founding
    \subsubsection{Shot Block Probability Model Validation}
    \label{sec:s_block_valid}

    Subsequently, for the theory-based shot block model, we compare the performance of different optimization methods: Powell, Nelder-Mead, CG, and SLSQP. These methods are commonly used in function optimization (see details in Section \ref{sec:rule_based_model}). The evaluation will be based on the binary Cross-Entropy Loss (CEL), where a lower CEL indicates better performance. The CEL is a commonly used scoring rule for probability estimation in a 2-class event.
    
    \begin{table}[]
    \caption{Theory-based shot block model optimization methods performance.}
    \label{tab:shotblock_opt}
    \begin{tabular}{lll}
    \toprule
    Optimization Methods & \multicolumn{1}{l}{Avg CEL} & \multicolumn{1}{l}{Std}\\
    \midrule
    \textbf{Powell} & \textbf{0.9220} & 0.2544            \\
    CG & 0.9229 & 0.2540                     \\
    L-BFGS-B & 0.9230 & 0.2540                     \\
    SLSQP & 0.9230 & 0.2540                     \\
    Nelder-Mead & 0.9345 & 0.2552     
    \\
    \botrule
    \end{tabular}
    \footnotetext{The table shows the average (Avg) CEL and standard deviation (Std) on the 5 cross-validations split validation set CEL. The table is ordered with the Avg CEL (the lower the better) in ascending order.}
    \end{table}

    Table \ref{tab:shotblock_opt} presents a comparison of the performance of various optimization methods for the theory-based shot block model. Among the optimization methods considered, the Powell method \citep{powell1964efficient} achieved the lowest CEL of 0.9220. Overall, all five optimization methods exhibited similar performance, indicating that the choice of optimization method had a minimal impact on the performance of the theory-based shot block model.

    \begin{table}[]
    \caption{The performance of shot block probability prediction models with machine learning and different feature sets.}
    \label{tab:shotblock}
    \begin{tabular}{lll}
    \toprule
    Model                 & \multicolumn{1}{l}{Avg CEL} & \multicolumn{1}{l}{Std} \\
    \midrule
    Ours (Proposed features)            & \textbf{0.4876}              & 0.0259                  \\
    ElasticNet            & 0.5417                       & 0.0077                  \\
      Ours (Basic shooter features only) & 0.5545  &  0.0168\\
     Ours (Unprocessed players features) & 0.5684 & 0.0081\\
    Historical percentage & 0.5783                       & 0.0011                  \\
    XGBoost               & 0.6354                       & 0.0372                  \\
    CatBoost              & 0.7096                       & 0.0272                       \\
    \botrule
    \end{tabular}
    \footnotetext{The proposed features are described in Section \ref{sec:xsot} and Figure \ref{fig:xsot_framwork}. The table shows the average (Avg) CEL and standard deviation (Std) on the 5 cross-validations split validation set CEL. The table is ordered with the Avg CEL (the lower the better) in ascending order.}
    \end{table}

    For the $DNN_{\text{block}}$ models, we assess their performance by comparing them against baseline models: historical percentage, ElasticNet, xGBoost, and CatBoost, using the same features set as the proposed model, as the above shot off model verification. The evaluation will be based on the binary Cross-Entropy Loss (CEL), where a lower CEL indicates better performance. The CEL is a commonly used scoring rule for probability estimation in a 2-class event.
    
    Table \ref{tab:shotblock} provides a summary of the performance comparison between models in estimating the probability of a shot being blocked. Our model, $DNN_{\text{block}}$, had outperformed all baseline models and achieved the lowest average CEL of 0.4876. This result validated that $DNN_{\text{block}}$ effectively provided inference for shot block probability and performed better than the baseline models.

    \subsubsection{Necessity of the Theory-Based Shot Block Model}
    \label{sec:necessity_tb}

    % \begin{table}[]
    % \caption{$DNN_{\text{block}}$ performance with different features sets.}
    % \label{tab:shotblock_feature}
    % \begin{tabular}{lll}
    % \toprule
    % Models                            & \multicolumn{1}{l}{Avg CEL}      & 
    % \multicolumn{1}{l}{Std}         \\
    % \midrule

    % \textbf{Proposed features} & \textbf{0.4876} & 0.0259\\
    % Basic shooter features only & 0.5545  &  0.0168\\
    % Unprocessed players features & 0.5684 & 0.0081\\%fill_nan
    % \botrule
    % \end{tabular}
    % \footnotetext{The proposed features are described in Section \ref{sec:xsot} and Figure \ref{fig:xsot_framwork}. The table shows the average (Avg) CEL and standard deviation (Std) on the 5 cross-validations split validation set CEL. The table is ordered with the Avg CEL (the lower the better) in ascending order}
    % \end{table}

    Additionally, we assess the necessity of the theory-based shot block model and compare the performance of the $DNN_{\text{block}}$ when fitted with different sets of features. Specifically, we consider the methodology features (details in Section \ref{sec:xsot}), an ablated version with only basic shooter features (details in Section \ref{sec:xsot}), advanced shooter features (details in Section \ref{sec:s_off_valid}), and direct utilization of non-shooter player's role and xy coordinates\footnote{An additional vector of size 22x4 was created to represent all 22 players, and their corresponding features (player role, location x, and location y, a teammate with the shooter or not (0 or 1)). This vector was concatenated with the basic shooter features from Section \ref{sec:xsot}, resulting in 22*4+5 features. For players who did not appear in the freeze frame data, we assigned a value of 0.} (Unprocessed player features).
    
    Furthermore, we verify the importance of combining the theory-based shot block and DNN models instead of using them independently. We compare their performance when used in combination and when used independently. The evaluation will be based on the binary cross-entropy loss (CEL), where a lower CEL indicates better performance. CEL is a commonly used scoring rule for probability estimation in a 2-class event.
    
    In Table \ref{tab:shotblock}, we demonstrated the necessity of the theory-based shot block model by comparing the use of different feature sets. The results indicate that the proposed shot block DNN model with the proposed features, utilized the theory-based shot block model's predicted shot block probability as features, achieved the best performance of 0.49. This provided evidence for the necessity of the theory-based shot block model in the framework.
    
    Finally, we validated the importance of combining the theory-based shot block and DNN models. From Table \ref{tab:shotblock_opt}, we observed that the theory-based shot block model alone achieved an average CEL of 0.92, and the $DNN_{\text{block}}$ alone achieved an average CEL of 0.55. However, when combined with the DNN model in the proposed method using the proposed features, as shown in Table \ref{tab:shotblock}, the average CEL largely improved to 0.49. This comparison highlighted the need for integrating both models, as it enhanced performance in estimating the probability of shot block.

    In summary, our analysis provided evidence of the effectiveness of $DNN_{\text{off}}$ and $DNN_{\text{block}}$ in estimating the probabilities of shot off target and shot block, respectively. Additionally, we validated the necessity of the theory-based shot block model and demonstrated the importance of combining it with $DNN_{\text{block}}$ to achieve improved performance.

    %features set direct use 0.5684 (fill nan) and common features only 0.5545(the need of rule-based)
    %compare DNN 0.4876 and rule-based 0.9220 (the need of combine rule-based and DNN)

    \subsection{Predicted Probability Validation}
    %weighted cost function for imbalance class

    \begin{table}[]
    \caption{Shot off prediction test set confusion matrix. }
    \label{tab:shotoff_cm}
    \begin{tabular}{lll}
    \toprule
             & Predicted 0   & Predicted 1   \\
    \midrule
    Actual 0 & $\bold{51.96\%}$ (159) & 48.04\% (147) \\
    Actual 1 & 46.89\% (98)  & $\bold{53.11\%}$ (111) \\
    \botrule
    \end{tabular}
    \end{table}

    \begin{table}[]
    \caption{Shot block prediction test set confusion matrix.}
    \label{tab:shotblock_cm}
    \begin{tabular}{lll}
    \toprule
             & Predicted 0   & Predicted 1   \\
    \midrule
    Actual 0 & $\bold{67.11\%}$ (203) & 32.89\% (100) \\
    Actual 1 & 32.60\% (36)  & $\bold{67.40\%}$ (74)  \\
    \botrule
    \end{tabular}
    \end{table}

    After verifying the models and framework, we proceeded to validate the predicted probabilities of shot off and shot block from the models with the test set. The $DNN_{\text{off}}$ and $DNN_{\text{block}}$ models were trained using inverse class weighted CEL. The model parameters were open-sourced and were applied for the analysis hereafter.

    The probabilities were then converted to binary values using a threshold of 0.5. The resulting confusion matrices for shot off and shot block could be found in Tables \ref{tab:shotoff_cm} and \ref{tab:shotblock_cm}, respectively. On average, the correctly assigned class had the highest probability. This indicates that the predictions made by the DNN models contained valuable information and were consistent with the observed outcomes.

    \subsection{xSOT and xOSOT Verification}

    Furthermore, to validate the proposed metrics, we calculated the total xSOT (expected Shot On Target), xOSOT (expected Offense Shot On Target), and an additional metric called $max\_prob = max(\text{xSOT, xOSOT})$, representing the maximum shot on probability a team could produce under a shot-taking situation. These calculations were performed for each team in the World Cup 2022, and averaged across matches (the final results were presented in Supplementary Table \ref{tab:world}).

    We employed the Pearson correlation metric to evaluate the information provided by the proposed metrics, existing metrics, and statistics due to the absence of ground truth data regarding the value of player actions and the probability of a shot being on target. The Pearson correlation enabled us to evaluate their respective relationships. This analysis helped determine which metrics aligned with each other and provided consistent insights.
    
    \begin{table}[]
    \caption{Correlation between the proposed metrics and the existing metrics.}
    \label{tab:correlation}
    \begin{tabular}{ll}
    \toprule
    Correlation Between & \multicolumn{1}{l}{Correlation Coefficient} \\
    \midrule
    Avg Goal, xG        & 0.46                                  \\
    Avg Goal, xSOT      & 0.58                                  \\
    xG, xSOT            & 0.88                                  \\
    xG, xOSOT           & 0.93                                  \\
    xG, max\_prob       & 0.95                                  \\
    \botrule
    \end{tabular}
    \footnotetext{Avg Goal (average goal), xG, xSOT, xOSOT, and max\_prob, represent the average match statistics of the teams involved in World Cup 2022.}
    \end{table}

    In Table \ref{tab:correlation}, we observed that the xSOT metric exhibited a higher correlation 0.58 with the average goal compared to the correlation between xG and the average goal (0.46). This suggested that xSOT was a better metric for approximating the final performance of a team in terms of goal scoring.

    Additionally, the proposed metrics, xSOT, xOSOT, and $max\_prob$, demonstrated high correlations with xG of 0.88, 0.93, and 0.95, respectively. This indicates that these metrics could effectively capture the attacking abilities of both teams and individual players, similar to how xG reflects the expected goal-scoring capability. Thus, the proposed metrics could provide valuable insights for evaluating the value of a player's action and the attacking prowess of teams and aligned with the established xG metric.

    \subsection{Optimial Strategy in World Cup 2022}

    %shot-taking situation with defender
    \begin{table}[]
    \caption{Payoff table for the attacker and closest defender.}
    \label{tab:payoff_table}
    \begin{tabular}{llll}
    \toprule
    &                            & \multicolumn{2}{c}{Closest Defender}                                              \\
    &                            & Blocking                            & Not Blocking                         \\ 
    \midrule                
    \multirow{2}{*}{Shooter} & \multicolumn{1}{c}{Shoot} & \multicolumn{1}{c}{$\overline{0.0866,-0.0866}$} & \multicolumn{1}{c}{$\underline{0.2508,-0.2508}$} \vspace{5pt}\\  
    & \multicolumn{1}{c}{Pass}  & \multicolumn{1}{c}{$\underline{\overline{0.2456,-0.2456}}$} & \multicolumn{1}{c}{0.2481,-0.2481} \\  
    \botrule
    \end{tabular}
    \footnotetext{The underline indicates the highest payoff for the attacker, given the defender selects the strategy in the column, and the overline indicates the highest payoff for the defender, given the attacker selects the strategy in the row. For more details regarding the calculation of Nash Equilibrium, please refer to \citep{tadelis2013game}. }
    \end{table}

    After successfully verifying the proposed models, metrics, and framework, we could now utilize them to uncover the optimal strategy for both the shooter and the closest defender in a shot-taking situation. By utilizing all available data, we filtered out situations where the set of filtered defenders $D$, with $|D|=0$, indicating no defender being considered in the baseline model; we were left with 1468 shot-taking situations for analysis. Since defenders in a blocking position had the option to either move out of the way (not blocking). On the other hand, it would be challenging to block the shot if the defender was not in a blocking position initially. 
    
    To determine the optimal strategy, we calculated the expected payoffs for each possible strategy profile and summarized them in Table \ref{tab:payoff_table}. According to the Nash equilibrium \citep{nash1951non}, the optimal strategy for the shooter was to pass the ball, while the optimal strategy for the closest defender was to block the shot. Deviating from this strategy would not yield a higher expected reward for either agent. Mixed strategies need not be considered since we had a pure strategy in this case. Moreover, with more data, the above analysis can be performed per team or even per player role as in \citep{tuyls2021game}.

    Furthermore, it is worth noting that the payoff difference between shooting and passing was significant ($\pm 0.15$) when the closest defender decided to block the shot. This suggested that, under expectation, there was an off-ball player who had a higher chance of successfully shooting on target. Therefore, passing became a more favorable option for the shooter, as it maximized the potential reward and could increase the team's chances of scoring.

    \subsection{EURO 2020 Shot-Taking Situations Analysis with xSOT and xOSOT}

    \begin{figure}[]
    \centering
    \includegraphics[width=0.7\textwidth]{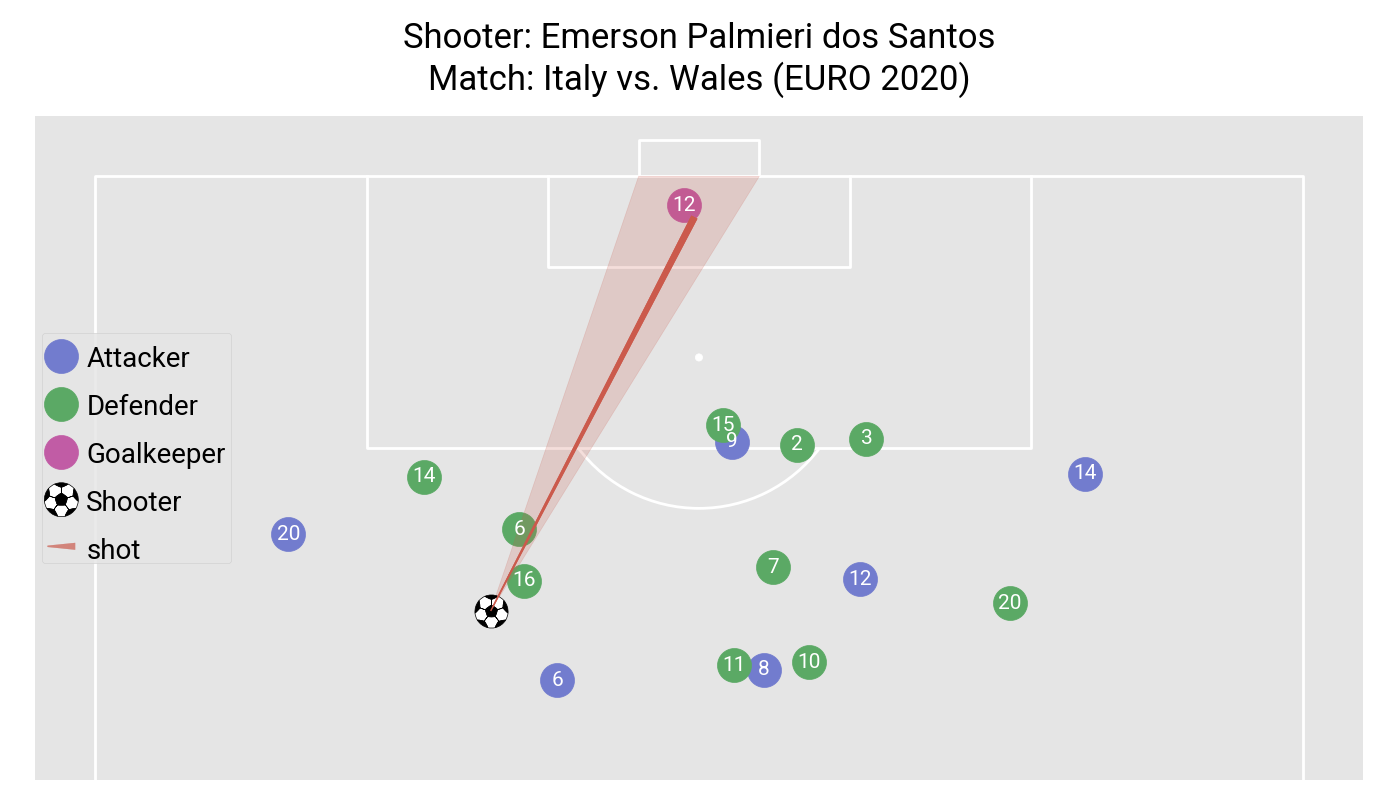}
    \caption{Shot-taking situation example image. The image included all players who appeared in the freeze frame from the match Italy vs. Wales, EURO 2020.}
    \label{fig:xsot_eg}
    \end{figure}
    
    As previously mentioned, it was expected that there was an off-ball attacking player (illustrated with the blue color dots in Figure \ref{fig:xsot_eg}) who had a higher chance of shooting on target in shot-taking situations. By utilizing xSOT and xOSOT, we could determine whether the shooter should take the shot or make a pass for the off-ball attacker to shoot (counterfactual), as well as identify the optimal recipient of the pass.

    Additionally, through the construction of xSOT and xOSOT, it became possible to estimate the probabilities of shot off, shot block, and control for each attacker involved in the situation. This information could help us understand why the off-ball attacker had a higher expected probability of shooting on target. By analyzing these probabilities, football players and analysts could gain insights into the positioning and other factors that contributed to the off-ball attacker's increased probability of shooting on target.
    
    \begin{table}[]
    \caption{Shot-taking situation example statistics.}
    \label{tab:xsot_stats}
    \begin{tabular}{lllll}
    \toprule
    \begin{tabular}[c]{@{}c@{}}Jersey\\ Number\end{tabular} & P(Shot On) & P(Shot Off) & P(Shot Block) & P(Control) \\
    \midrule
    9                                                       & 0.27 & 0.32        & 0.22          & 0.59       \\
    20                                                      & 0.23 & 0.60        & 0.03          & 0.63       \\
    14                                                      & 0.17 & 0.67        & 0.16          & 0.99       \\
    12                                                      & 0.15 & 0.63        & 0.20          & 0.89       \\
    6                                                       & 0.05 & 0.53        & 0.18          & 0.17       \\
    Shooter                                                 & 0.03 & 0.51        & 0.46          & -          \\
    8                                                       & 0.00 & 0.61        & 0.40          & 0.60      
    \\
    \botrule
    \end{tabular}
    \footnotetext{Ordered by P(Shot On) descendingly. The P(Shot On) for the shooter is estimated with the xSOT value, and for off-ball attacker a is estimated with xOSOT without the maximization function, P(Shot Off) and P(Shot Block) are the estimations from DNNs, and P(Control), are estimated with PPCF.}
    \end{table}

    Figure \ref{fig:xsot_eg} illustrates a shot-taking situation from the EURO 2020 match between Italy and Wales. Table \ref{tab:xsot_stats} provides the values of the proposed metrics for each attacker involved in the freeze frame. In this scenario, Attacker 9 (Jersey number) exhibited the highest probability of shooting on target 0.27 and the lowest probability of shooting off target 0.32. Since Attacker 9 is closer to the goal line. On the other hand, Attacker 20 had the second-best probability of shooting on target 0.23, and the lowest probability of the shot being blocked by defenders 0.03, because Attacker 20 faced fewer defenders but was in a further position to the goalline. Furthermore, Attacker 14 demonstrated the highest probability of controlling the ball 0.99, as no defenders were around. Therefore, passing to Attacker 14 would be the optimal choice to maintain possession and increase the team's chances of retaining control of the ball. 
    
    By analyzing these metrics, we could gain valuable insights into the shooting, blocking, and controlling probabilities of each attacker, which could guide decision-making in shot-taking situations and enhance the team's overall performance.

\section{Conclusion}
\label{sec:conclusion}
    In summary, this research aims to provide an effective and data-driven method to comprehensively analyze the interaction strategy between the shooter and defender. To achieve this objective, we have proposed a novel framework that integrates the use of machine learning, a theory-based approach, and game theory. We have validated the models $DNN_{\text{off}}$ and $DNN_{\text{block}}$ for estimating event outcomes, the metrics xSOT and xOSOT for valuing players' actions, and provided examples to analyze team strategies and shot-taking situations with open-access data. We expect this framework to help teams gain a more in-depth understanding of shot-taking situations. Specifically, in difficult or controversial situations, xSOT would help perform an objective analysis, ultimately enhancing teams' performance.
    
    In the future, since the metric xSOT provides the expected probability for all players in the data, the skill level of each player would affect the probabilities in the metric. It would be possible to estimate team or player-specific xSOT by incorporating player skills-related features into the DNN models, as demonstrated in \cite{yeung2023framework}. Additionally, we assumed the interaction was a static one-stage game due to the lack of velocity and other detailed data. If velocity and other detailed data become available, it would be possible to define a multi-stage game that incorporates the expected movement of the players. In conclusion, with more data related to players, shot-taking situations, and football matches, a more comprehensive version of this framework could be developed. Nevertheless, we expect that this framework will serve as inspiration for analyzing complex interaction situations, particularly in the realm of sports.

\backmatter

\bmhead{Acknowledgments}
This work was financially supported by JST SPRING, Grant Number JPMJSP2125. The author, Calvin C. K. Yeung, would like to take this opportunity to thank the “Interdisciplinary Frontier Next-Generation Researcher Program of the Tokai Higher Education and Research System.” 

\section*{Declarations}
\begin{itemize}
\item Funding: This work was financially supported by JST SPRING, Grant Number JPMJSP2125.
\item Competing interests: The authors have no competing interests to declare that are relevant to the content of this article.
\item Ethics approval: Not applicable
\item Consent to participate: Not applicable
\item Consent for publication: Not applicable
\item Availability of data and materials: The data of the research is publicly available with details in \url{https://github.com/statsbomb/statsbombpy}

\item Code availability: The code for the model is available at
% \ifarxiv
% \url{https://github.com/calvinyeungck/Football-Match-Event-Forecast}.
% \else
\url{https://github.com/calvinyeungck/Analyzing-Two-Agents-Interaction-in-Football-Shot-Taking-Situations}.
% \fi
\item Authors' contributions: All authors contributed to the study conception and design. Data preparation, modeling, and analysis were performed by Calvin C. K. Yeung. The first draft of the manuscript was written by Calvin C. K. Yeung and all authors commented on previous versions of the manuscript. All authors read and approved the final manuscript.
\end{itemize}

\begin{appendices}
\setcounter{table}{10}
\setcounter{figure}{5}
\setcounter{equation}{10}

 \section{Shot Outcome}
    This section summarizes how the shot outcomes are grouped. Details of each shot outcome and the corresponding group are provided in Supplementary Table \ref{tab:shot_outcome_grouping}.
    \label{sec:shot_outcome_appendix}

    \begin{table}[]
    \caption{Shot outcome grouping.}
    \label{tab:shot_outcome_grouping}
    \begin{tabular}{ll}
    \toprule
    Group           & Shot Outcome                                                                     \\
    \midrule
    Shot On Target  & Goal                                                                             \\
                    & Saved: Saved by goal keeper                                                      \\
    Shot Off Target & Off T: off target                                                                \\
                    & Wayward: An unthreatening shot                                                   \\
                    & Post: A shot that hit one of the three posts                                     \\
                    & Saved Off Target: A shot that was saved by the goalkeeper but was not on target. \\
    Shot Block      & Blocked: Blocked by defender                                                     \\
    Removed         & Saved to Post: Goalkeeper saves the shot and bounces off the goal frame \\
    \botrule
    \end{tabular}
    \footnotetext{Further details available on \url{https://github.com/statsbomb/open-data}.}
    \end{table}
    
    \section{Data Independence Test}
    \label{sec:data_independence_test}
    It is important to understand if the outcome of the previous shots affects the following shots. Since it shows if the interaction between the shooter and the defender is a static or dynamic game.
    However, given that we defined the shot outcome as three discrete classes, most common statistical tests have certain limitations on testing the sequential independence of the shot outcome.

    In tests for discrete outcomes, the Runs Test \citep{gibbons2020nonparametric} is applicable for two classes only, and the Runs Up and Down Test\citep{gibbons2020nonparametric} is for ordered outcome only. Moreover, the Chi-Square Test and Markov chain-based test ignore the order of the events. Lastly, the Ljung-Box test, Durbin-Watson test, Autocorrelation, or Time series based test requires distribution assumption, and for continuous variables only.
    
    Nonetheless, we perform the Chi-Square Test for Independence (ne previous outcome), the hypothesis as follows:
    \begin{itemize}
        \item Null Hypothesis (H0): The outcomes in the sequence are independent of each other.
        \item Alternative Hypothesis (H1): The outcomes in the sequence are dependent on each other.
    \end{itemize}
    
    \begin{table}[]
    \caption{Contingency table for the shot outcome.}
    \label{tab:contingency}
    \begin{tabular}{lllll}
    \toprule
    \multicolumn{1}{l}{}              & \multicolumn{1}{l}{} & \multicolumn{3}{c}{Following Outcome} \\
    \multicolumn{1}{l}{}              & Grouped Outcome      & Shot Off   & Shot On   & Shot Block   \\
    \midrule
    \multirow{3}{*}{Previous Outcome} & Shot Off             & 427        & 341       & 275          \\
                                      & Shot On              & 343        & 286       & 220          \\
                                      & Shot Block           & 273        & 222       & 187  \\
                                      \botrule
    \end{tabular}
    \end{table}
    Supplementary Table \ref{tab:contingency} was the contingency table for shots outcome, with 4 Degrees of freedom. The chi-square test gave a test statistic of 0.6163 and a p-value of 0.9612. Since the p-value was greater than 0.05, the null hypothesis was not rejected, therefore, the current outcome was independent of the previous outcome.
    However, this did not imply that the outcomes were sequentially independent, and therefore, we had to assume sequential independence (Static game).

    \section{Hyperparmeters}
    This section presents the details and best grid searched values of the hyperparameters for the DNN models and the parameters value of the theory-based shot block model. For DNN models and theory-based shot block models, the specific details can be found in Supplementary Tables \ref{tab:proposed} and \ref{tab:rule_based_opt_param} respectively.
    \label{sec:hyperparameters}
    \begin{table}[]
    \caption{Proposed models hyperparameters.}
    \label{tab:proposed}
    \begin{tabular}{llll}
    \toprule
    Hyperparameters & Grid Searched Value        & Shot Off Model & Shot Block Model \\
    \midrule
    num\_layers     & 1,2,3                      & 1              & 2                \\
    hidden\_dim     & 32,64,128                  & 128            & 64               \\
    dropout\_rate   & 0.0,0.1,0.2                & 0              & 0                \\
    activation      &\begin{tabular}[c]{@{}c@{}}nn.ReLU,\\ nn.Sigmoid,\\ nn.Tanh\end{tabular} & nn.Tanh        & nn.Sigmoid       \\
    embedding1      & 1,2,3                      & 2              & 1                                       \\
    \botrule
    \end{tabular}
    \footnotetext{The columns Shot Off Model and Shot Block Model indicate the best hyperparameters for the respective model.}
    \end{table}
    
    Where the explanation for each hyperparameter for the DNN model is as follows:
    \begin{itemize}
    \item num$\_$layer: Numbers of hidden layers in the DNN.
    \item hidden$\_$dim: Numbers of nodes in each hidden layer.
    \item dropout$\_$rate: Dropout rate for each hidden layer.
    \item activation: Activation function for each hidden node.
    \item embedding1: embedding layer output dimension, for embedding the position feature.
    \end{itemize}
    
    \begin{table}[]
    \caption{Theory-based shot block model optimized parameter.}
    \label{tab:rule_based_opt_param}
    \begin{tabular}{ll}
    \toprule
    Parameters & Optimize value \\
    \midrule
    c\_1       & 36.9463        \\
    c\_2       & 12.3579        \\
    c\_3       & 0.4998         \\
    c\_4       & 0.1577         \\
    a          & -2.3098      \\
    \botrule
    \end{tabular}
    \end{table}

    \section{OBSO and PPCF}
    The off-ball scoring opportunity (OBSO) \citep{spearman2018beyond} was modeled using the following equation:
    \label{sec:ppcf}
    \begin{equation}
    P(G|D)=\sum_{r\in\mathbb{R}\times\mathbb{R}}P(G_r|C_r,T_r,D)P(C_r|T_r,D)P(T_r|D)
    \end{equation}
    where,
    \begin{itemize}
        \item $D$ represents the current game stats.
        \item $G_r$ represents a goal scored from location r.
        \item $C_r$ represents the passing team controls the ball at location r.
        \item $T_r$ represents the next event that happens at location r.
    \end{itemize}

    The potential pitch control field (PPCF) model is represented by the second term $P(C_r|T_r,D)$ in the aforementioned equation. The equation for PPCF is as follows:

    \begin{align}
    \label{eq12}
    \frac{dPPCF_j}{dT}(t,\overrightarrow{r},T|s,\lambda_j)&=(1-\sum_k PPCF_k(t,\overrightarrow{r},T|s,\lambda_j))f_j(t,\overrightarrow{r},T|s)\lambda_j \nonumber\\
    f_j(t,\overrightarrow{r},T|s)&=\bigg[1+e^{-\pi\frac{T-\tau_{exp}(t,\overrightarrow{r})}{\sqrt{3}s}}\bigg]^{-1}
    \end{align}

    \noindent where $f_j(t,\overrightarrow{r},T|s)$ denotes the probability that player $j$ will reach location $\overrightarrow{r}$ and control the football within time $T$. $\lambda_j$ and $s$ are optimizable parameters, and $\tau_{exp}(t,\overrightarrow{r})$ is the expected interception time, calculated based on the player's initial location, acceleration, and maximum speed.

    In this study, the PPCF is calculated by integrating Supplementary Equation \ref{eq12} from time 0 to time $T$, where $T$ represents the travel time required from the shooter's location to the off-ball attacker.

    \section{Would Cup 2022 Statistics}
    This section provides detailed statistics and metrics of teams in the World Cup 2022. The specific details can be found in Supplementary Table \ref{tab:world}.

    \begin{table}[]
    \caption{World Cup 2022 team statistics.}
    \label{tab:world}
    \begin{tabular}{lllllll}
    \toprule
    Team          & Placement       & \multicolumn{1}{l}{xSOT} & \multicolumn{1}{l}{xOSOT} & \multicolumn{1}{l}{Max Prob} & \multicolumn{1}{l}{Avg Goal} & \multicolumn{1}{l}{xG} \\
    \midrule
    Argentina     & Final           & 2.47                     & 3.07                      & 4.05                          & 2.14                              & 1.76                   \\
    Australia     & Round of 16     & 1.38                     & 1.83                      & 2.43                          & 1.00                              & 0.74                   \\
    Belgium       & Group Stage     & 1.75                     & 2.29                      & 3.27                          & 0.33                              & 1.37                   \\
    Brazil        & Quarter-finals  & 2.91                     & 4.57                      & 5.67                          & 1.60                              & 2.44                   \\
    Cameroon      & Group Stage     & 1.96                     & 2.22                      & 3.16                          & 1.33                              & 1.37                   \\
    Canada        & Group Stage     & 1.90                      & 2.87                      & 3.77                          & 0.67                              & 1.31                   \\
    Costa Rica    & Group Stage     & 0.91                     & 0.72                      & 1.25                          & 1.00                              & 0.56                   \\
    Croatia       & 3rd Place Final & 1.86                     & 2.73                      & 3.32                          & 1.14                              & 1.40                    \\
    Denmark       & Group Stage     & 2.35                     & 2.83                      & 3.80                          & 0.33                              & 1.62                   \\
    Ecuador       & Group Stage     & 1.56                     & 2.43                      & 3.11                          & 1.33                              & 1.33                   \\
    England       & Quarter-finals  & 2.17                     & 2.48                      & 3.55                          & 2.60                              & 1.77                   \\
    France        & Final           & 2.32                     & 3.34                      & 4.29                          & 2.29                              & 1.91                   \\
    Germany       & Group Stage     & 3.75                     & 5.78                      & 7.13                          & 2.00                              & 2.87                   \\
    Ghana         & Group Stage     & 1.67                     & 2.00                         & 2.68                          & 1.67                              & 1.04                   \\
    Iran          & Group Stage     & 1.78                     & 2.46                      & 3.26                          & 1.33                              & 1.32                   \\
    Japan         & Round of 16     & 1.81                     & 2.83                      & 3.37                          & 1.25                              & 1.11                   \\
    Mexico        & Group Stage     & 1.84                     & 3.21                      & 3.83                          & 0.67                              & 1.78                   \\
    Morocco       & 3rd Place Final & 1.40                      & 2.11                      & 2.62                          & 0.86                              & 1.00                      \\
    Netherlands   & Quarter-finals  & 1.76                     & 1.66                      & 2.58                          & 2.00                              & 1.06                   \\
    Poland        & Round of 16     & 1.39                     & 1.82                      & 2.25                          & 0.75                              & 0.78                   \\
    Portugal      & Quarter-finals  & 2.35                     & 3.33                      & 4.25                          & 2.40                              & 1.56                   \\
    Qatar         & Group Stage     & 1.05                     & 1.86                      & 2.07                          & 0.33                              & 0.70                    \\
    Saudi Arabia  & Group Stage     & 1.86                     & 2.10                       & 3.03                          & 1.00                              & 1.27                   \\
    Senegal       & Round of 16     & 1.85                     & 3.04                      & 3.68                          & 1.25                              & 1.55                   \\
    Serbia        & Group Stage     & 2.50                      & 2.87                      & 3.93                          & 1.67                              & 1.32                   \\
    South Korea   & Round of 16     & 1.54                     & 2.94                      & 3.52                          & 1.25                              & 1.72                   \\
    Spain         & Round of 16     & 2.90                      & 3.36                      & 4.29                          & 2.25                              & 1.70                    \\
    Switzerland   & Round of 16     & 1.60                      & 1.89                      & 2.54                          & 1.25                              & 1.04                   \\
    Tunisia       & Group Stage     & 1.47                     & 2.48                      & 3.00                          & 0.33                              & 1.24                   \\
    United States & Round of 16     & 1.98                     & 2.78                      & 3.56                          & 0.75                              & 1.50                    \\
    Uruguay       & Group Stage     & 2.04                     & 3.02                      & 3.61                          & 0.67                              & 1.41                   \\
    Wales         & Group Stage     & 1.05                     & 2.13                      & 2.58                          & 0.33                              & 0.95                  \\
    \botrule
    \end{tabular}
    \end{table}




\end{appendices}

\newpage
\bibliography{sn-bibliography}% common bib file
% \input output.bbl

\end{document}